\documentclass[letterpaper, 10 pt, conference]{ieeeconf}

\IEEEoverridecommandlockouts
\usepackage{graphicx}
\usepackage{amsmath}
\usepackage{amssymb}
\usepackage{booktabs}
\usepackage{array}
\usepackage{xcolor}
\usepackage{url}
\usepackage[hidelinks]{hyperref}

\newcommand{\pai}{$\pi_0$}
\newcommand{\paih}{$\pi_{0.5}$}

\definecolor{good}{RGB}{23,114,69}
\definecolor{bad}{RGB}{154,44,44}

\definecolor{instrbg}{RGB}{231,225,243}
\newlength{\instrW}\newlength{\instrH}\newlength{\instrTmp}
\newsavebox{\instrSave}
\newcommand{\instrtext}[2]{\footnotesize\centering ``#1''\\(Item \emph{mass}: #2)}
\newcommand{\instrheight}[5]{
  \sbox{\instrSave}{\parbox[t]{\instrW}{\instrtext{#2}{#3}}}%
  \setlength{#1}{\dimexpr\ht\instrSave+\dp\instrSave\relax}%
  \sbox{\instrSave}{\parbox[t]{\instrW}{\instrtext{#4}{#5}}}%
  \setlength{\instrTmp}{\dimexpr\ht\instrSave+\dp\instrSave\relax}%
  \ifdim\instrTmp>#1\setlength{#1}{\instrTmp}\fi}
\newcommand{\instrbox}[3]{
  \colorbox{instrbg}{\parbox[t][#1][t]{\instrW}{\instrtext{#2}{#3}}}}

\title{\LARGE \bf
Do VLAs Understand and Adapt to the Objects They Handle, or Simply Replay Learned Behaviors?}

\author{Xinnuo Xu\\
\texttt{xuxinnuo@yahoo.com}%
\thanks{preprint}}

\begin{document}
\maketitle
\thispagestyle{empty}
\pagestyle{empty}

\begin{abstract}
\looseness=-1 This paper asks whether VLA generalization is grounded in a global understanding of objects' physical properties that enables policies to adapt their motion to unseen setups, or if policies simply replay the motions they've learnt that happen to succeed in new setups. The former reflects genuine generalization; the latter reflects incidental robustness. We first examine awareness of physical properties in seven VLAs by applying linear probing and representational similarity analysis (RSA) to their activations. We find that physical properties, including \emph{mass}, \emph{fragility}, \emph{deformability}, \emph{friction} and \emph{size} are less decodable than non-physical properties such as \emph{semantic category}, \emph{material}, \emph{sound} and \emph{price} in nearly every modality stream. Compared with their base VLMs, robot pre-training weakens the linear encoding of physical properties in the language stream. Neither pre-training nor downstream fine-tuning strengthens the alignment between physical-property differences and activation distances. We then ask whether the weak physical information present in these activations shapes the actions a VLA generates. In a controlled LIBERO case study, we increase the \emph{mass} of an in-domain object and signal the change through language or vision. Most VLAs use similar lifting behaviour for the heavier and original-\emph{mass} objects, leading to task success declines. The few exceptions change their behaviour in response to lexical or visual cues rather than to \emph{mass} itself. These results suggest that VLAs encode physical properties weakly and do not reliably use them to adapt their motion.

\end{abstract}

\section{Introduction}

\begin{figure}[t]
\centering
\setlength{\tabcolsep}{2pt}
\setlength{\instrW}{\dimexpr0.48\columnwidth-2\fboxsep\relax}
\instrheight{\instrH}{turn on the stove and put the moka pot on it}{68.2\,g}{turn on the stove and put the moka pot \textbf{full of coffee} on it}{648.3\,g}
\newlength{\instrHmug}
\instrheight{\instrHmug}{put the white mug on the plate and put the chocolate pudding to the right of the plate}{16.5\,g}{put the white mug on the plate and put the chocolate pudding to the right of the plate}{186.1\,g}
\begin{tabular}{@{}cc@{}}
\textbf{Original} & \textbf{Full} \\
\includegraphics[width=0.48\columnwidth]{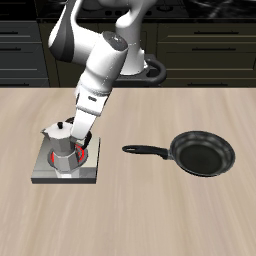} &
\includegraphics[width=0.48\columnwidth]{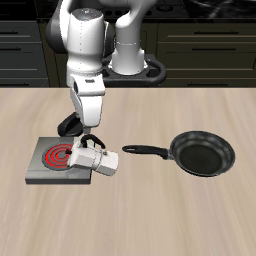} \\
\instrbox{\instrH}{turn on the stove and put the moka pot on it}{68.2\,g} &
\instrbox{\instrH}{turn on the stove and put the moka pot \textbf{full of coffee} on it}{648.3\,g} \\ \noalign{\vskip 8pt}
\includegraphics[width=0.48\columnwidth]{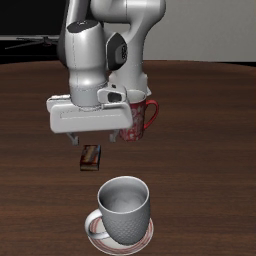} &
\includegraphics[width=0.48\columnwidth]{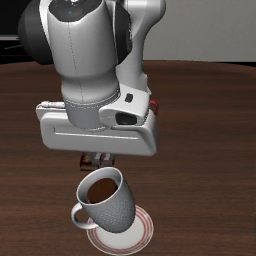} \\
\instrbox{\instrHmug}{put the white mug on the plate and put the chocolate pudding to the right of the plate}{16.5\,g} &
\instrbox{\instrHmug}{put the white mug on the plate and put the chocolate pudding to the right of the plate}{186.1\,g} \\
\end{tabular}
\caption{\textbf{Signalling a heavier object changes how VLAs act.} Top: SmolVLA, moka pot; the instruction names the pot full of coffee and its \emph{mass} is increased, and the pot ends tilted on the stove. Bottom: $\pi_{0.5}$, white mug; the instruction is unchanged, the mug is rendered full of coffee and its \emph{mass} is increased, and the mug drops from the gripper. Left: original task, where both succeed.}
\label{fig:teaser}
\vspace{-10pt}
\end{figure}

Vision Language Action models (VLAs) trained on large robot corpora show some robustness to unseen objects and environments~\cite{liberopro,liberoplus,liberox,liberopara,quest,liberogen}.
A fundamental question, however, remains: is this generalization grounded in a global understanding of object physical properties, e.g. \emph{mass} and \emph{deformability}, that lets the policy adapt its motion to new setups, or does it just replay the motions they've learnt in training, and assume they are robust in unseen setups? We examine two dimensions.

\textbf{Awareness}: do VLAs represent the physical properties of the objects they manipulate in their latent representations? Following~\cite{vlaprobesym,vlaprobeval,featobs,rsa,rsanlp}, we quantify awareness with two metrics. \emph{Decodability} fits a linear probe to the model's activations and asks whether a property can be read out of them; \emph{representation geometry} uses representational similarity analysis (RSA) to ask whether pairwise distances between object activations track differences in that property, such that objects with similar property values are represented similarly.
Across seven widely used VLAs on the OXE and DROID datasets, we find that all physical properties, including \emph{mass}, \emph{fragility}, \emph{deformability}, \emph{friction} and \emph{size}, are less decodable than the other properties, such as \emph{category}, \emph{material}, \emph{sound} and \emph{price}, in almost every modality stream, and no stream's pairwise activation distances align most strongly with differences in a physical property. Moreover, physical properties are less decodable from the language stream after robot-interaction pre-training and downstream fine-tuning than in the corresponding base VLMs, while neither stage strengthens the alignment between physical-property differences and activation distances (Sec.~\ref{sec:experiments}).

\textbf{Behavioral adaptation}: do VLAs adapt their actions to changes in objects'
physical properties? We answer this with a controlled case study of \emph{mass} in
LIBERO (Sec.~\ref{case_study}). As shown in Fig.~\ref{fig:teaser}, we select manipulable containers from LIBERO, preserve their geometry while adding the \emph{mass} of realistic contents, and signal the change through language (top row in Fig.~\ref{fig:teaser}) or vision (bottom row in Fig.~\ref{fig:teaser}). Six VLAs execute $23$ tasks with \emph{mass}-amended and original objects under otherwise identical setups. Success falls for \emph{mass}-amended objects, since for most VLAs the lifting-phase actions deviate no further from the original. Three policies are exceptions, but none tracks \emph{mass} by exhibiting the human pattern of lifting heavier objects more slowly. Instead, the action divergence arises from either lexical rewording or pixel-level changes.

As a conclusion, the results suggest that VLAs encode physical properties weakly and do not reliably use them to adapt their motion.
\section{Related Work}

\textbf{Probing VLA representations.} Linear probes are widely used in LLMs, where
affine readouts recover high-level concepts encoded as activation
directions~\cite{linrep,catgeom}. Applied to VLAs, they reveal symbolic object and action
states~\cite{vlaprobesym}, value-like success structure~\cite{vlaprobeval}, and
observable end-effector variables such as yaw and gripper
aperture~\cite{featobs}. Richer techniques expose complementary structure; for example,
sparse autoencoders recover motion primitives, task semantics and episode-specific
features~\cite{sae}. We use linear probes for their simple, uniform readout across
architectures.
Prior work shows that robot training can degrade broad multimodal capabilities and
pretrained representations, with retention depending on the action representation and
training recipe~\cite{vlm2vla,preserverep,vlatrace}. We study this degradation through
the lens of \emph{object} representations, asking which physical properties are retained
or lost; our analysis is therefore complementary rather than a rediscovery of the
broader effect.

\textbf{Representational similarity analysis.} Introduced in systems
neuroscience~\cite{rsa}, RSA relates two systems through their dissimilarity structures
and is a standard tool for connecting neural and symbolic representations in language
models~\cite{rsanlp}. We apply it to VLAs to ask what a probe cannot: whether a property
organizes the representation. A property can be linearly decodable yet contribute
nothing to the distances between object embeddings.

\textbf{Object physical properties.} PhysObjects~\cite{physobjects} annotates
\emph{mass}, \emph{fragility}, \emph{deformability}, \emph{material} and other
properties in household-object images, and
uses them to ground a VLM that improves object selection by a high-level LLM planner.
Human-norm resources provide reference labels for some properties, including measured
\emph{mass} in YCB~\cite{ycb} and \emph{size} in
THINGSplus~\cite{thingsplus}; we use these to
validate our annotation (Sec.~\ref{sec:annotation}). That line of work targets \emph{task
planning}, i.e. which object to pick for a goal, whereas we target \emph{motion
execution}: whether a policy's low-level actions reflect the properties of the object
it holds.

\section{Dataset and Annotation}\label{sec:annotation}

In this paper we augment three widely-used datasets: OXE~\cite{oxe}, DROID~\cite{droid}, and LIBERO~\cite{libero}, with commonsense object-property annotations elicited from GPT-4.1. We restrict our study to the \emph{pick-and-place} task family, and from each dataset select the one-hand-manipulable objects: $200$ for OXE, $88$ for DROID, and $13$ for LIBERO (e.g.\ \emph{apple}, \emph{mug}, \emph{moka pot}). For each object we annotate twelve properties spanning physical mechanics, semantic identity and everyday context: \emph{mass}, \emph{fragility}, \emph{deformability}, \emph{friction}, \emph{size}, \emph{material}, \emph{sound}\footnote{Whether the object is characteristically associated with producing a sound, e.g.\ a bell or blender vs.\ a towel.}, \emph{price}, \emph{category}\footnote{The object's superordinate semantic category, e.g.\ \emph{food}, \emph{tool}}, \emph{location}\footnote{The location where the object is typically found, e.g.\ \emph{kitchen} or \emph{office}.}, \emph{colour}\footnote{The object's most characteristic colour.}, and \emph{disposability}\footnote{Whether the object is typically single-use and discarded after use (e.g.\ a napkin) rather than reused (e.g.\ a mug).}.
We refer to the first five as \emph{physical properties}: by common sense, they directly affect how an object is grasped, lifted and placed.
Among these, \emph{mass}, \emph{fragility}, \emph{deformability}, \emph{friction}, \emph{size}, \emph{sound} and \emph{price} are ordinal, and \emph{material}, \emph{category}, \emph{location}, \emph{colour} and \emph{disposability} are categorical.
For a categorical property we query GPT-4.1 one object at a time by name, e.g.\ ``In what type of room is a \{object\} most typically found?'' for \emph{location}, and record its chosen label from a fixed label set.
For an ordinal property we instead pose pairwise comparisons by name, ``Which is more X, \{object$_1$\} or \{object$_2$\}?'', over every within-dataset pair (each asked in both orders to cancel order bias). We then fit a Bradley-Terry (BT) model~\cite{bt} to the resulting win matrix to obtain \textbf{a per-object score and a ranking within each dataset}.

\textbf{Annotation validity.} We assess GPT-4.1's annotations against human-labelled
reference datasets. For the categorical properties we report classification accuracy
and Cohen's $\kappa$ against a human gold label: \emph{category}
($0.83$, $\kappa\,0.81$; THINGSplus~\cite{thingsplus}) and \emph{material}
($0.88$, $\kappa\,0.85$; PhysObjects~\cite{physobjects}). For the ordinal properties
we report the balanced accuracy of GPT-4.1's pairwise ``which is more X?'' judgments
against the human-norm ordering: \emph{size} $0.97$ (THINGSplus~\cite{thingsplus}),
\emph{sound} $0.96$ (Amsel et al.~\cite{amsel}), \emph{mass} $\sim\!0.84$ (YCB
measured \emph{mass}~\cite{ycb}), and \emph{fragility} $0.81$ and \emph{deformability}
$0.79$ against the PhysObjects \cite{physobjects}.
Every accuracy above meets or exceeds the $\ge\!70\%$ inter-human-annotator agreement that PhysObjects
requires to accept a human gold label~\cite{physobjects}, so we treat the GPT-4.1
labels as trustworthy. The remaining properties, \emph{friction}, \emph{price},
\emph{location}, \emph{colour} and \emph{disposability}, have no established human
dataset and are annotated by GPT-4.1 without validation.

\textbf{\emph{Mass} in LIBERO.} LIBERO's simulated \emph{mass} does not reflect an
object's real-world weight. Every object is assigned a uniform density
($100\,\mathrm{kg\,m^{-3}}$) with no explicit \emph{mass}, so MuJoCo
computes each object's \emph{mass} as density $\times$ collision-mesh volume, i.e. a
pure \emph{size} proxy that ignores \emph{material} and contents. Because the LIBERO
fine-tunes were trained under this physics, for LIBERO we discard the GPT-4.1 real-world
\emph{mass} annotation and label \emph{mass} with the simulator's density-derived value
instead.

\section{Models and Methodology}\label{sec:methodology}
We study seven widely used VLA policies spanning the three dominant action-decoding paradigms: autoregressive discrete tokens ($\pi_0$-FAST~\cite{pi0fast}, OpenVLA~\cite{openvla}), flow-matching action experts (\pai~\cite{pi0}, \paih~\cite{pi05}, SmolVLA~\cite{smolvla}, GR00T-N1.7~\cite{groot,grootn17}), and JEPA-style latent prediction (VLA-JEPA~\cite{vlajepa}). We ask whether these models \emph{internally represent} object physical properties, rather than merely mapping pixels or text to actions. Borrowing from mechanistic interpretability in LLMs, we study the geometry of the frozen models' layer-wise activations through linear probing and representational similarity analysis (RSA), asking where and how those latent representations encode each object property (property details in Sec.~\ref{sec:annotation}).

\begin{figure}[!t]
\centering
\includegraphics[width=\columnwidth]{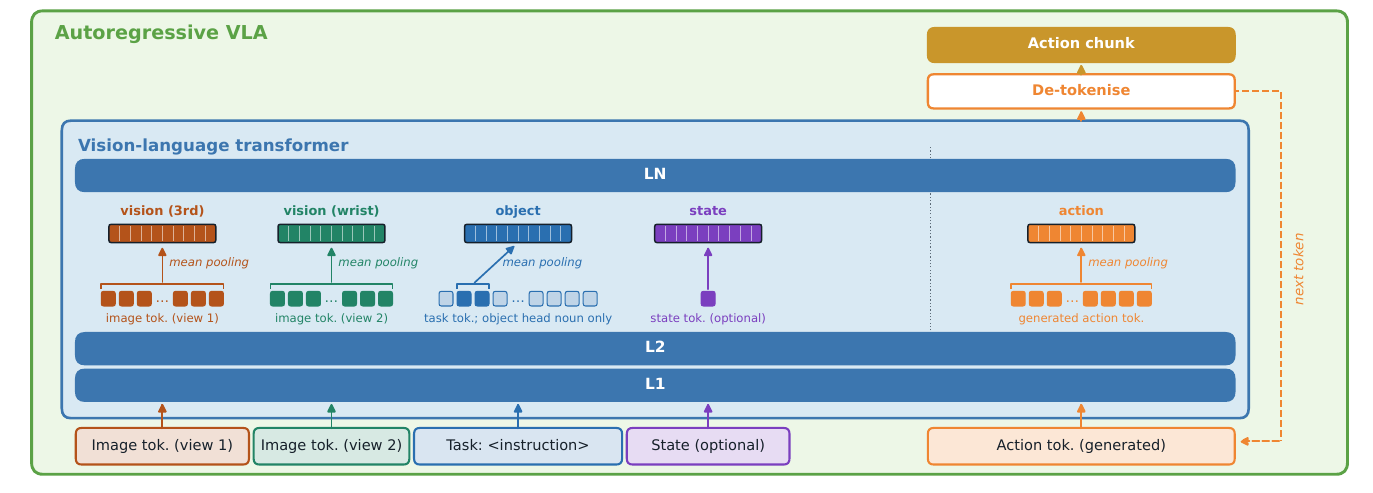}\\[3pt]
\includegraphics[width=\columnwidth]{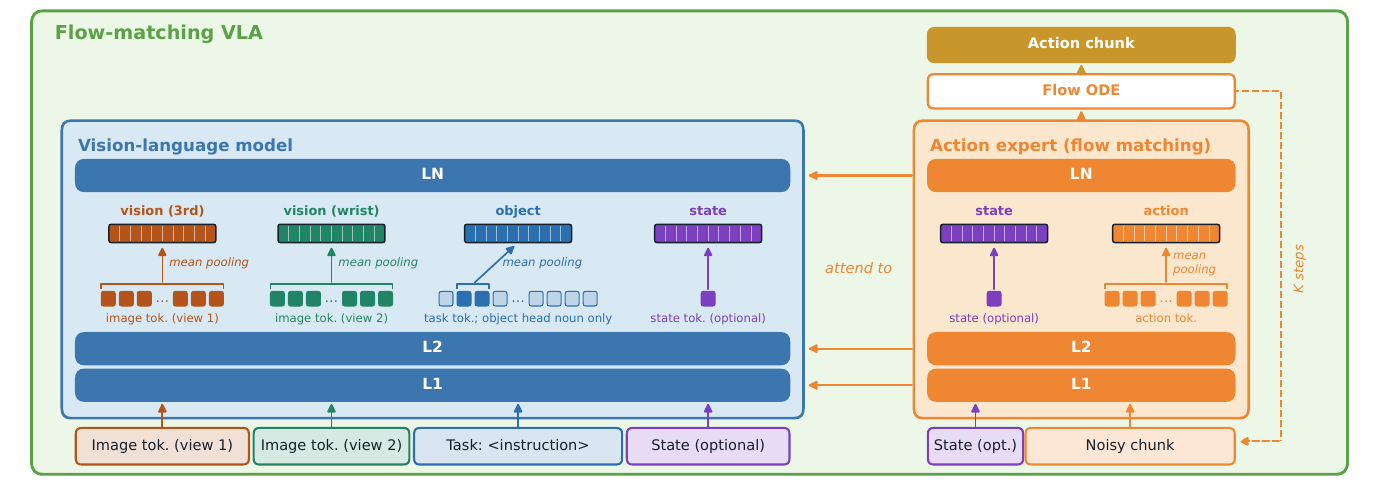}
\vspace{-15pt}
\caption{\textbf{The two VLA families we study, and how we extract their latent representation.} \emph{Top:} an \textbf{autoregressive} policy (OpenVLA, \pai-FAST) runs one transformer over a multimodal prefix and generates discrete action tokens one at a time. \emph{Bottom:} a \textbf{flow-matching} policy (\pai, \paih, VLA-JEPA, SmolVLA, GR00T-N1.7) attends the VLM prefix, layer by layer, from a separate action expert that denoises a noisy action chunk over $K$ flow steps. In both, we record the activations at \emph{every} layer and mean-pool a modality's token span into one embedding: per-view \emph{vision}, the \emph{object} head-noun text tokens, the optional \emph{state} token, and the \emph{action} tokens (generated tokens, or the expert's tokens at the \textit{last} flow step). The input modalities and their ordering vary across models. For clarity we omit VLA-JEPA's world-model latent tokens and separate embodied-action tokens from this figure. See Sec.~\ref{sec:methodology}.}
\label{fig:arch}
\vspace{-20pt}
\end{figure}

\textbf{Latent representation extraction.} Fig.~\ref{fig:arch} shows how we extract frozen latent representations from the two mainstream VLA architecture classes, autoregressive (top) and flow-matching (bottom). Consider a frame $t$ of a trajectory in which the robot manipulates object $o$. Given the task instruction, the camera views and the robot state at $t$, we run a single forward pass and record the activations at \emph{every} layer $\ell$. For each modality $m$, we then mean-pool the activations over that modality's token span into one vector $x^{(m)}_{o,\ell}\in\mathbb{R}^{d}$: \emph{object} pools the instruction's object head-noun tokens, \emph{vision} pools the patch tokens of each camera view, \emph{state} takes the robot-state token where one exists, and \emph{action} pools the generated action tokens (autoregressive) or the action-expert tokens at the last flow step (flow-matching). We call each pooled span a \emph{modality stream}: object-text (the object's head-noun tokens), the third-person camera and the wrist camera (each view is its own stream), robot state, and action tokens.

\textbf{Physical Property Decodability -- Probing} answers whether a physical property is linearly readable from a latent representation \cite{linrep,catgeom,featobs,vlaprobesym,vlaprobeval}. A \textbf{\emph{probe}} is an affine map applied to a frozen representation. For a categorical property with classes $c\in\{1,\dots,K\}$ ($K\ge2$), we define the probe as a linear map $g(x^{(m)}_{o,\ell}) = W x^{(m)}_{o,\ell}+b$, with $W\in\mathbb{R}^{K\times d}$ and $b\in\mathbb{R}^{K}$, which predicts the class with the highest score, $\hat{y}_o=\arg\max_c g(x^{(m)}_{o,\ell})_c$. We train it as a multinomial logistic regression, minimising the $\ell_2$-regularised cross-entropy over the training objects $\mathcal{O}_{\text{tr}}$ with label $y_o$:
\begin{equation}
    \small
\mathcal{L}=-\sum_{o\in\mathcal{O}_{\text{tr}}}\log\mathrm{softmax}\big(g(x^{(m)}_{o,\ell})\big)_{y_o}+\gamma\lVert W\rVert_F^2 ,
\end{equation}

For an \emph{ordinal} property, each object $o$ is annotated with a continuous within-dataset Bradley-Terry (BT) score $s_o$, so we probe \emph{order}. For an object pair $(i,j)$ whose scores are separated by a margin, $|s_i-s_j|>\epsilon$ with $\epsilon=0.05$, we define the probe as a linear map on the embedding difference, $h_{ij} = h(x^{(m)}_{i,\ell}, x^{(m)}_{j,\ell}) = w^{\top}(x^{(m)}_{i,\ell}-x^{(m)}_{j,\ell})+b$, with $w\in\mathbb{R}^{d}$ and $b\in\mathbb{R}$, which predicts that $i$ ranks above $j$ when $h>0$. With label $z_{ij}=\mathbf{1}[s_i>s_j]$, we train it as a binary logistic regression, minimising the $\ell_2$-regularised cross-entropy over the training pairs $\mathcal{P}_{\text{tr}}$:
\begin{equation}
    \small
\mathcal{L}=-\!\!\sum_{(i,j)\in\mathcal{P}_{\text{tr}}}\!\Big[z_{ij}\log\sigma(h_{ij})+(1-z_{ij})\log\big(1-\sigma(h_{ij})\big)\Big]+\gamma\lVert w\rVert_2^2 ,
\end{equation}
where $\sigma$ is the logistic sigmoid. Every training pair enters in both orders, $(i,j)$ and $(j,i)$, so the two labels are balanced.

The chance levels of the categorical and ordinal probes are $1/K$ and $0.5$, respectively. We evaluate each probe by its balanced accuracy on held-out test sets $\mathcal{O}_{\text{test}}$ and $\mathcal{P}_{\text{test}}$, where $\mathcal{O}_{\text{tr}}\cap\mathcal{O}_{\text{test}}=\varnothing$ and $\mathcal{P}_{\text{tr}}\cap\mathcal{P}_{\text{test}}=\varnothing$. A property is decodable from modality $m$ at layer $\ell$ when this accuracy is above chance, and the closer it is to $1$, the more strongly the property is linearly encoded.

\textbf{Representation's geometry -- RSA.} Representational Similarity Analysis (RSA) reveals whether a property \emph{organizes} a representation's geometry. For the $N$ objects of a dataset we build two Representational Dissimilarity Matrices (RDMs). The neural RDM uses the cosine distance between pooled embeddings,
\begin{equation}
    \small
D^{\text{rep}}_{ij}=1-\frac{\langle x_i,x_j\rangle}{\lVert x_i\rVert\,\lVert x_j\rVert},
\end{equation}
and the property RDM uses the gap in BT score for ordinal properties,
$D^{\text{prop}}_{ij}=|s_i-s_j|$, or label disagreement for categorical ones,
$D^{\text{prop}}_{ij}=\mathbf{1}[y_i\neq y_j]$. The RSA score is the Spearman rank
correlation between the upper triangles of the two RDMs,
\begin{equation}
    \small
\mathrm{RSA}=\rho_s\big(\{D^{\text{rep}}_{ij}\}_{i<j},\,\{D^{\text{prop}}_{ij}\}_{i<j}\big).
\end{equation}

$RSA \in [-1, 1]$. A high score means that, objects with similar property values have embeddings that are closer than objects far apart on the property. A score near $0$ means embedding distances are unrelated to the property. A below $0$ means objects with similar property values sit farther apart.

Unlike the probe, RSA fits no parameters. a property can be decodable along one direction yet have RSA near $0$, if other properties dominate the distances between embeddings. 

\section{Experiment Results}\label{sec:experiments}

This section reports results aggregated across models, while Sec.~\ref{case_study}
examines individual models in detail.

\begin{figure}[!t]
\centering
\includegraphics[width=\linewidth]{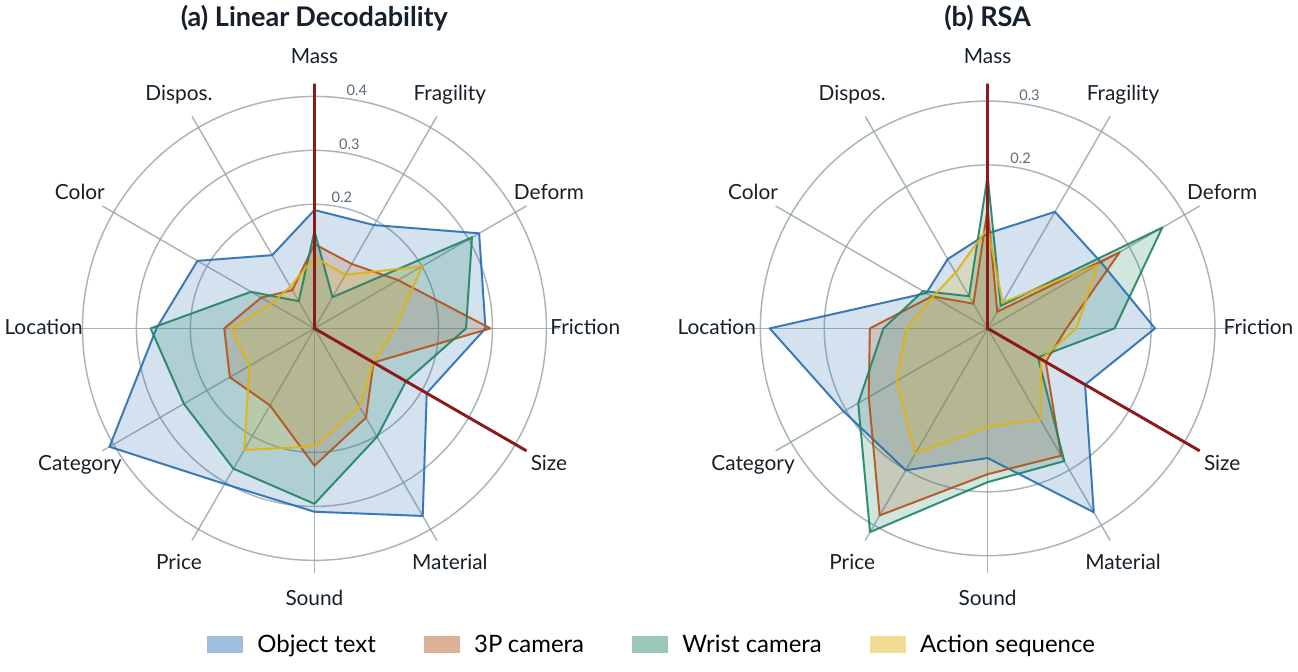}
\vspace{-20pt}
\caption{\textbf{Physical properties by modality stream in pre-trained VLAs.} \textbf{(a)}~Decodability: the gap between probe accuracy and \emph{chance} ($1/K$ for categorical properties, $0.5$ for ordinal ones; Sec.~\ref{subsec:pre-train-vla-decodability}). \textbf{(b)}~Representation geometry: RSA (Sec.~\ref{subsec:pre-train-vla-dominate}). Both metrics are defined in Sec.~\ref{sec:methodology}. Each polygon is one modality stream and each axis is one property. The dark-red boundary axes bracket the physical properties (\emph{mass} to \emph{size}). Values are averaged over the seven pre-trained VLAs (six for the wrist camera) and over OXE, DROID.}
\label{fig:streams}\label{fig:rsa_streams}
\vspace{-15pt}
\end{figure}

\begin{figure*}[!t]
\centering
\includegraphics[width=\textwidth]{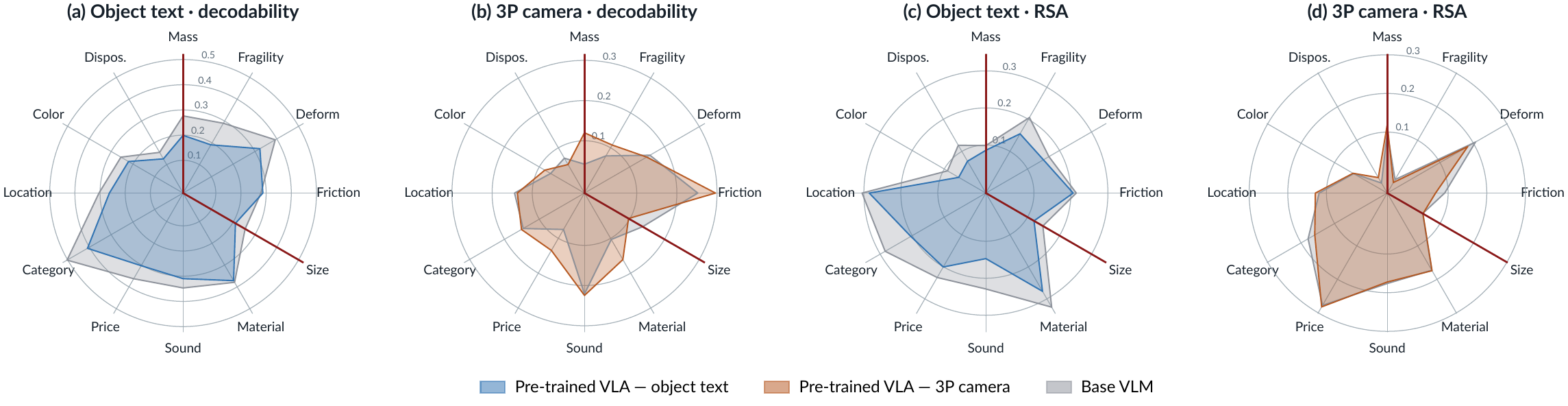}
\vspace{-20pt}
\caption{\textbf{Base VLM vs. pre-trained VLA, by modality stream}. Each
radar overlays the base VLM (grey) and its pre-trained VLA (blue for the object-text stream, orange for the third-person camera stream) over the twelve properties, with the physical properties bracketed by the dark-red axes. \textbf{(a,\,b)}~Decodability, i.e.\ probe accuracy minus chance; \textbf{(c,\,d)}~Representation geometry, measured by RSA. Both are averaged over all VLA--VLM pairs and over OXE, DROID. See Sec.~\ref{subsec:pre-train-vla-vlm} for the discussion.}
\label{fig:delta_vlm}\label{fig:rsa_delta_vlm}
\end{figure*}

\subsection{Property Decodability in Pre-trained VLAs}\label{subsec:pre-train-vla-decodability}

In this section, we ask where physical properties are decodable in pre-trained VLAs.

\textbf{Setup.} We probe the pre-trained checkpoint of each VLA: \pai{} (\texttt{pi0\_base}), \paih{} (\texttt{pi05\_base}), $\pi_0$-FAST (\texttt{pi0fast-base}), SmolVLA (\texttt{smolvla\_base}), VLA-JEPA (\texttt{VLA-JEPA-Pretrain}), GR00T-N1.7-3B and OpenVLA-7B, on two real-robot demonstration datasets. 

From OXE, we sample one episode for each of the $200$ annotated objects. From DROID, we collect $410$ episodes covering the annotated objects, with a median of 4 episodes per object. 
From each episode, we take the frame at the \emph{early} interaction phase ($30\%$ of the way through) from every available camera: the third-person view only on OXE, and both the third-person and wrist views on DROID. The instruction names the object in a pick-and-place template, ``pick up the $\langle$obj$\rangle$''. No robot state is used: \pai, SmolVLA, GR00T-N1.7 receive a zero state vector, and \paih{} and $\pi_0$-FAST receive the instruction without state text. 
Each model then generates actions conditioned on the language instruction and the camera view(s).

For each dataset, we hold out $20\%$ of the objects for testing. Ordinal test pairs are formed only among the held-out objects ($780$ pairs on OXE and $105$ on DROID), and training pairs are subsampled to $400$ per dataset. We train an independent probe for every layer and every modality stream of a model, and report the layer that performs best under 2-fold cross-validation on the training objects.

\textbf{Findings.} Fig.~\ref{fig:streams}(a) reports decoding accuracy \emph{minus chance}, averaged over datasets and then over the seven models; the dark-red boundary axes bracket the physical properties (\emph{mass} to \emph{size}). The wrist stream averages six models on DROID only, as OXE dataset and OpenVLA model take a single image.
We find that (i)~On average, every one of the 12 properties is decodable above chance from each of the four modality streams; (ii)~Object-text is the most decodable stream for 10 of the 12 properties, and action tokens are the least decodable stream for 8 of 12; and (iii)~The physical properties are \emph{not} the most decodable: from object-text, \emph{mass} ($+0.19$) and \emph{fragility} ($+0.19$) rank 11th and 10th of 12, well below \emph{category} ($+0.41$), \emph{material} ($+0.37$), \emph{sound} ($+0.31$) and \emph{price} ($+0.30$). The same trend holds for the other modality streams.

\begin{figure*}[!t]
\centering
\includegraphics[width=\textwidth]{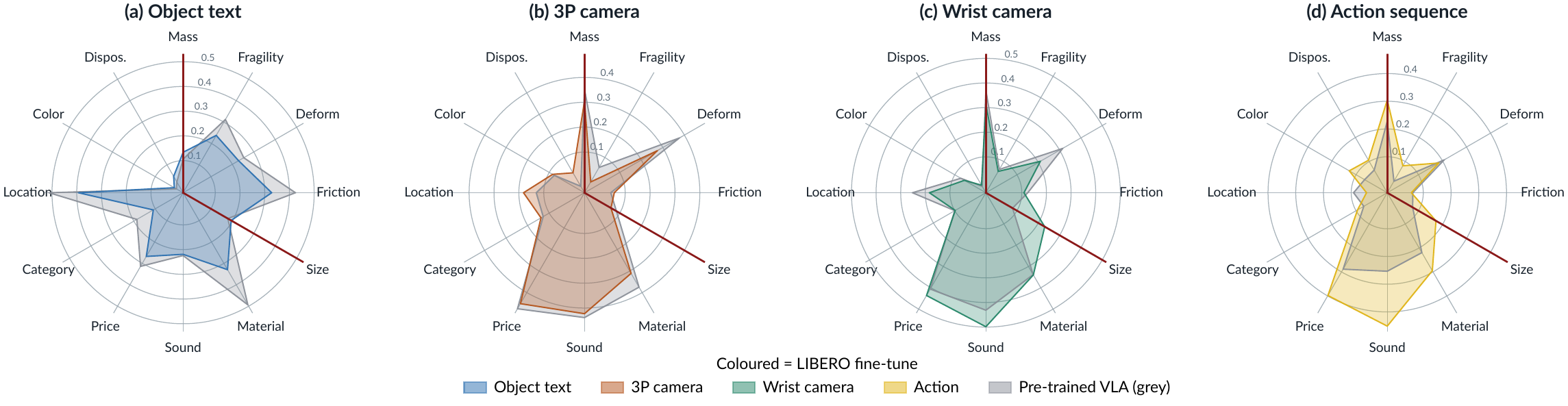}
\vspace{-10pt}
\caption{\textbf{RSA: LIBERO fine-tuning versus pre-training}, one radar per
modality stream. Each panel overlays the pre-trained VLA (grey) and its LIBERO fine-tune
(in the stream's colour) over the twelve properties (clockwise, Sec.~\ref{sec:annotation}
order; physical properties bracketed by the dark-red axes), averaged over the seven
families with a LIBERO checkpoint. LIBERO \emph{mass} is simulator \emph{mass};
categorical RSA is defined on LIBERO, but its labels are nearly single-class
(\emph{location}: 14 of 15 objects
\emph{kitchen}). Discussed in Sec.~\ref{subsec:pre-train-vla-ft-vla}.}
\label{fig:rsa_delta_libero}
\vspace{-5pt}
\end{figure*}

\subsection{Representation Geometry of Pre-trained VLAs}\label{subsec:pre-train-vla-dominate}

Here, we ask which properties organize the representation geometry of pre-trained VLAs.

\textbf{Setup.} We compute RSA on the same early-frame representations, models and datasets as Sec.~\ref{subsec:pre-train-vla-decodability}. RSA fits no parameters, so it uses all annotated objects of a dataset without a train/test split; for each modality stream we report the peak over layers, which is an upper bound. Fig.~\ref{fig:streams}(b) aggregates over all tested VLAs.

\textbf{Findings.} (i)~The geometry of object-text (blue) is organized most by two non-physical properties, \emph{location} ($0.29$) and \emph{material} ($0.28$), and less by any physical property ($\le 0.21$). (ii)~The two vision streams (orange and green) and the action stream (yellow) share a common profile: they are organized most by \emph{price}, a non-physical property (third-person $0.28$, wrist $0.31$, action $0.17$), and next by \emph{deformability}, a physical one ($0.18$, $0.26$, $0.14$). \emph{Mass}, \emph{friction} and \emph{size} contribute moderately, while \emph{fragility} leaves no trace in their geometry ($\le 0$), even though it is decodable from every stream (Fig.~\ref{fig:streams}(a)). Overall, non-physical properties are more prominent than physical ones in the representation geometry of every stream, and the strongest organizer of each stream is a non-physical property.

\subsection{Base VLMs vs.\ Pre-trained VLAs}\label{subsec:pre-train-vla-vlm}

Here, we ask how robot pre-training, i.e.\ further training a VLM on physical interaction data, changes the decodability and representation geometry of physical properties relative to the other object properties.

\textbf{Setup.} We compare each VLA with the VLM it is built on: \pai, \paih{} and $\pi_0$-FAST with PaliGemma-3B~\cite{paligemma}; SmolVLA with SmolVLM2-500M~\cite{smolvlm}; VLA-JEPA with Qwen3-VL-2B~\cite{qwen3vl}; GR00T-N1.7 with Cosmos-Reason2~\cite{cosmosreason2}; and OpenVLA with Prismatic~\cite{prismatic}. 
Each VLM receives the same pick-and-place instruction and third-person frame as its VLA, in a single forward pass without generation.
Datasets, splits and layer selection follow Sec.~\ref{subsec:pre-train-vla-decodability}.

\textbf{Findings.} Fig.~\ref{fig:delta_vlm} overlays each VLM and its VLA, averaged over all VLA--VLM pairs. (a)~Robot pre-training lowers object-text decodability for 11 of the 12 properties, and physical properties are hit hardest: three of the four largest drops are \emph{fragility} ($-0.10$), \emph{mass} ($-0.08$) and \emph{deformability} ($-0.07$). (b)~Vision decodability changes little overall ($+0.02$ on average across properties), yet its largest gain is physical: \emph{mass} rises by $+0.08$ and \emph{fragility} by $+0.03$. (c,\,d)~The properties that most strongly organize the representation stay largely the same after robot pre-training: the RSA ranking of the 12 properties correlates between VLMs and VLAs (Spearman $\rho=0.91$ for object-text and $0.97$ for vision). Object-text RSA falls for all 12 properties, but physical properties lose about half as much as the others (mean $-0.02$ vs.\ $-0.05$), whereas vision RSA changes by at most $0.024$. Training on physical interaction data thus weakens the linear encoding of physical properties in language and slightly strengthens \emph{mass} and \emph{friction} in vision, but it does not make physical properties more prominent in the representation geometry: their RSA does not increase in either stream, and non-physical properties such as \emph{location}, \emph{material} and \emph{price} remain the strongest organizers.

\subsection{Pre-trained VLAs vs.\ Downstream Dataset FT}\label{subsec:pre-train-vla-ft-vla}

\looseness=-1 Here, we ask how downstream fine-tuning (FT), i.e.\ further training a pre-trained VLA on the physical interaction data of a downstream dataset, changes the decodability and representation geometry of physical properties relative to the other object properties, evaluated on the objects of that same dataset.

\textbf{Setup.} We compare the pre-trained checkpoints of five VLAs from Sec.~\ref{subsec:pre-train-vla-decodability}---$\pi_0$-FAST, SmolVLA, VLA-JEPA, GR00T-N1.7 and OpenVLA---with their corresponding released LIBERO fine-tunes. We use 8 LIBERO demonstration episodes for each of the $13$ objects ($104$ episodes in total). As in Sec.~\ref{subsec:pre-train-vla-decodability}, we take the frame $30\%$ of the way through each episode from both cameras, feed no robot state, and let each model generate its actions. To keep the input in-domain, the instruction is the episode's own LIBERO instruction. With only $13$ objects, probe estimates are too noisy, so we report only RSA. Each object's embedding is averaged over its 8 episodes before computing RSA, and layer selection follows Sec.~\ref{subsec:pre-train-vla-dominate}.

\textbf{Findings.} Fig.~\ref{fig:rsa_delta_libero} shows that in object-text (blue), RSA falls much less for physical than for non-physical properties (mean $-0.01$ vs.\ $-0.07$), with the largest drops in \emph{material} ($-0.14$), \emph{sound} ($-0.09$) and \emph{price} ($-0.09$).
In the vision streams (red and green), physical properties instead lose RSA while the others gain (third-person $-0.04$ vs.\ $+0.03$; wrist $-0.03$ vs.\ $+0.03$). In the action stream (yellow), 7 of the 12 properties gain RSA and the largest gain is non-physical (\emph{sound}, $+0.13$). As with robot pre-training, fine-tuning on LIBERO fails to make physical properties more prominent in the representation geometry.

\section{LIBERO Case Study: \emph{Mass}} \label{case_study}

\begin{figure*}[!t]
\centering
\begin{tabular}{@{}c@{\hspace{4pt}}c@{}}
\includegraphics[height=123pt]{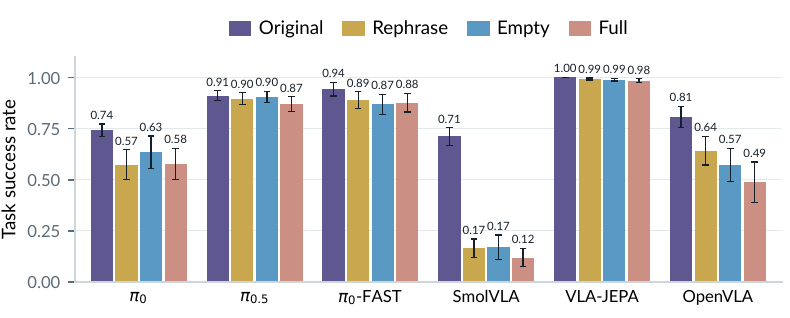} &
\includegraphics[height=123pt]{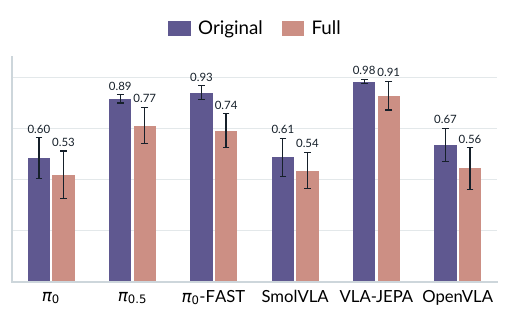} \\
\footnotesize (a) Language-based cue & \footnotesize (b) Vision-based cue \\
\end{tabular}
\caption{\textbf{Task success rate under mass-relevant cues.} Success rate averaged over tasks (13 with the language-based cue, 10 with the vision-based cue; 20 episodes per task and setup), with error bars showing the standard error across tasks. \textbf{(a)}~The four language-based setups are listed in Table~\ref{tab:milk_conditions}. \textbf{(b)}~The two vision-based setups are illustrated in the bottom row of Fig.~\ref{fig:teaser}. Setups are detailed in Sec.~\ref{subsec:mass_setup} and results discussed in Sec.~\ref{subsec:mass_success}.}
\label{fig:libero_success}
\end{figure*}

Sec.~\ref{subsec:pre-train-vla-decodability} showed that mass is linearly decodable from every modality stream of every VLA we study. Decodability, however, does not mean the policy uses the information. In this section, we ask whether a cue that signals the change of object mass in language or vision, actually changes how a VLA acts.

\subsection{Controlled Experimental Setup}\label{subsec:mass_setup}

LIBERO assigns every object the same default density of $100$\,kg/m$^3$, so an object's simulated mass is proportional to its collision volume and reflects its size rather than its weight. For containers, this default mass ($5$--$68$\,g) is close to that of an empty container. To isolate mass, we keep each object's geometry and initial pose fixed and change only its mass and the cue that signals it. We take the containers that appear in the training data of the four LIBERO suites and split them by whether their contents are visible, so that mass is signalled to the policy through language alone for one group and through vision alone for the other.

\looseness=-1 \textbf{Language-based cue.} This group holds sealed, opaque containers: \textit{alphabet soup}, \textit{cream cheese}, \textit{salad dressing}, \textit{BBQ sauce}, \textit{ketchup}, \textit{tomato sauce}, \textit{butter}, \textit{milk}, \textit{chocolate pudding} and \textit{orange juice} (LIBERO-Object), the \textit{moka pot} (LIBERO-10) and the \textit{wine bottle} (LIBERO-Goal, in two tasks), 13 tasks in total. Filling these containers changes no pixel, so mass can be signalled only through the instruction. Table~\ref{tab:milk_conditions} shows the setups derived from each \textit{Original} LIBERO instruction. \textit{Empty} names the container's default state explicitly (``empty carton of milk'') at the default mass. \textit{Full} names the filled state (``full carton of milk'') and adds the mass of the contents, computed from their density and $85\%$ of the object's collision volume, with the rendering unchanged. As a control, \textit{Rephrase} uses the same restructured name without the mass-indication word (``carton of milk''). It shows whether a change in the policy's behavior comes from rewording the object name or from the mass the word implies.

\textbf{Vision-based cue.} This group holds open containers: two types of \textit{mugs} (LIBERO-10, three tasks) and a \textit{bowl} (LIBERO-Goal and LIBERO-Spatial, seven tasks), 10 tasks in total. The instruction is never changed; \emph{Full} setup renders visible contents (coffee or soup) inside the container and adds their mass, computed from their density and rendered volume. The contents have no collision. This group has only the original and full conditions. The bottom row in Fig~\ref{fig:teaser} shows an example. Full masses range from $48$ to $648$\,g, $8$--$25\times$ the default. 

\begin{table}[t]
\centering
\setlength{\tabcolsep}{4pt}
\resizebox{\columnwidth}{!}{%
\begin{tabular}{@{}llr@{}}
\toprule
\textbf{Setup} & \textbf{Instruction} & \textbf{Mass} \\
\midrule
Orig & pick up the \textbf{milk} and place it in the basket & $31.7$\,g \\ \hline
Reph & pick up the \textbf{carton of milk} and place it in the basket & $31.7$\,g \\\hline
Empty    & pick up the \textbf{empty carton of milk} and place it in the basket & $31.7$\,g \\
Full     & pick up the \textbf{full carton of milk} and place it in the basket & $308.9$\,g \\\hline
\bottomrule
\end{tabular}}
\par\vspace{5pt}
\refstepcounter{table}\label{tab:milk_conditions}
\parbox{\columnwidth}{Table~\thetable: \textbf{Setups of the language-based cue, shown for the \textit{milk} task.} Bold: the object phrase, the only part of the instruction that changes. \textit{Original}, \textit{Rephrase} and \textit{Empty} keep the carton's default LIBERO mass; \textit{Full} adds $277.2$\,g of milk content, computed from its density ($1.03$\,g/cm$^3$) and $85\%$ of the carton's collision volume. Setup details are in Sec.~\ref{subsec:mass_setup}.}
\end{table}

\begin{table}[t]
\centering
\setlength{\tabcolsep}{4pt}
\begin{tabular}{@{}lcccccc@{}}
\toprule
& \pai & \paih & $\pi_0$-FAST & SmolVLA & VLA-JEPA & OpenVLA \\
\midrule
\multicolumn{7}{@{}l}{\emph{Language-based cue}, $z$ action} \\
Rephrase & 1.01 & 0.97 & 1.02 & \textbf{1.40} & 0.99 & 1.02 \\
Empty    & 1.00 & 0.98 & 1.03 & \textbf{1.35} & 1.01 & 1.06 \\
Full     & 1.01 & 0.99 & 1.03 & \textbf{1.38} & 1.01 & 1.05 \\
\midrule
\multicolumn{7}{@{}l}{\emph{Language-based cue}, full action space (7-D)} \\
Rephrase & 1.01 & 0.98 & 1.01 & \textbf{1.27} & 1.00 & 1.01 \\
Empty    & 1.00 & 0.99 & 1.02 & \textbf{1.19} & 1.02 & 1.04 \\
Full     & 1.01 & 0.99 & 1.03 & \textbf{1.21} & 1.01 & 1.03 \\
\midrule
\midrule
\multicolumn{7}{@{}l}{\emph{Vision-based cue}, $z$ action} \\
Full     & 0.98 & 1.07 & \textbf{1.33} & 0.98 & 1.15 & 1.03 \\
\midrule
\multicolumn{7}{@{}l}{\emph{Vision-based cue}, full action space (7-D)} \\
Full     & 0.99 & 1.01 & \textbf{1.25} & 0.98 & 1.10 & 1.02 \\
\bottomrule
\end{tabular}
\par\vspace{5pt}
\refstepcounter{table}\label{tab:grasp_z_agreement}
\parbox{\columnwidth}{Table~\thetable: \textbf{Agreement of commanded actions right after the grasp.} $R$ compares episodes generated under \textit{Original} setup with those generated under the setup named in the row: the mean distance between the two setups divided by the mean distance within each setup. $R\approx1$ means the mass cue moves the commanded actions no further than run-to-run variation within \textit{Original}. Bold: the episodes from the two setups differ in at least half of the tasks. Discussed in Sec.~\ref{subsec:mass_episode}.}
\end{table}

\subsection{Findings: Task Success}\label{subsec:mass_success}

Each model executes every task for $20$ episodes per setup, starting from the same initial states in every task. An episode counts as successful when LIBERO's built-in goal check is satisfied. Fig.~\ref{fig:libero_success} reports the success rate averaged over tasks (13 language-based, 10 vision-based), with error bars showing the standard error across tasks. 

With the language-based cue (Fig.~\ref{fig:libero_success}a), \textit{Full} lowers success for every model, most for SmolVLA ($0.71\rightarrow0.12$) and OpenVLA ($0.81\rightarrow0.49$). However, rewording the object name at the default mass, in either \textit{Rephrase} or \textit{Empty}, costs nearly as much, and adding the mass in \textit{Full} lowers success by at most a further $0.08$ relative to \textit{Empty} for any model. \paih{} and VLA-JEPA stay within $0.05$ of \textit{Original} in every language setup.
With the vision-based cue (Fig.~\ref{fig:libero_success}b), where the instruction is unchanged, \textit{Full} again lowers success for every model, most for $\pi_0$-FAST ($0.93\rightarrow0.74$) and \paih{} ($0.90\rightarrow0.77$). \paih{}, $\pi_0$-FAST and VLA-JEPA lose more success when the heavier object is signalled only through vision than through language, whereas \pai{}, SmolVLA and OpenVLA lose less.

\subsection{Episode Analysis}\label{subsec:mass_episode}

Success rate measures how robust a model is to a mass change signalled through language or vision; VLA-JEPA, for example, loses less than $0.10$ under both cues. Success alone, however, cannot tell whether a model stays robust because it adjusts its actions to the heavier object, or because it issues the same actions as for the original object and the added mass is simply not enough to make them fail. We therefore analyse the episodes the models generate.

\looseness=-1 Human prehension studies show that mass shows up in how an object is lifted rather than in how it is approached~\cite{weir1991}. Heavier objects are lifted later and with a lower lift velocity~\cite{eastough2007}. 
A policy that uses mass should likewise reveal it right after the grasp, when the object first loads the arm. 
We therefore locate the \emph{grasp point} in each episode as the step at which the gripper opening first drops below $90\%$ of its initial width.\footnote{When an episode closes the gripper more than once, we take the closure that immediately precedes the object's rise.} 
We then take the $10$ consecutive commanded actions starting at the \emph{grasp point} in every episode,\footnote{For the models that plan and execute $50$-step action chunks (\pai, \paih{} and SmolVLA), we specifically keep only episodes whose $10$ steps lie inside the chunk that contains the \emph{grasp point}. A heavier object makes the arm track its commands less closely, so a chunk planned after the grasp is conditioned on different camera views and robot state; a difference in its actions could then reflect this changed input rather than the model's knowledge of mass.} and test whether VLAs command a different lift in this window than they do for the original object.

Let $a^{(e)}_t\in\mathbb{R}^{D}$ be the action commanded at step $t$ of episode window $e$, where $D$ is the number of action dimensions. In LIBERO, $D=7$. The distance between two windows $e$ and $e'$ is the root mean square (RMS) difference of their actions,
\begin{equation}
\small
d(e,e')=\sqrt{\frac{1}{TD}\sum_{t=1}^{T}\big\lVert a^{(e)}_t-a^{(e')}_t\big\rVert_2^2 }.
\end{equation}
where $T=10$ is the window length. We use RMS rather than a scale-invariant measure such as cosine similarity or correlation because the magnitude of an action is exactly what we want to compare: lifting twice as fast doubles the action while leaving its direction unchanged, so cosine and correlation stay the same, whereas the RMS distance doubles.

For a task, let $\mathcal{E}_O$ and $\mathcal{E}_X$ be the episode windows of \textit{Original} and of a changed setup $X$. We average $d$ over all pairs within $\mathcal{E}_O$ ($\bar d_{OO}$), within $\mathcal{E}_X$ ($\bar d_{XX}$) and across the two sets ($\bar d_{OX}$), and compare the cross-setup distance with the run-to-run variation within each setup: $R=\frac{\bar d_{OX}}{\tfrac12\big(\bar d_{OO}+\bar d_{XX}\big)}$.
We average $R$ over tasks, and report the results in Table~\ref{tab:grasp_z_agreement}.

We report $R$ over the full action space ($D=7$) and over the vertical component $\Delta z$ of the commanded end-effector displacement alone ($D=1$), the direction in which the load acts. Both give the same picture. Most entries sit at $R\approx1$: changing the setup moves a model's actions no further than simply repeating \textit{Original} does. These models never adapt their lift to the cue, and instead rely on one policy being robust enough across masses.

\looseness=-1 Three models are exceptions. However, none of them adapts in a way that tracks mass. SmolVLA departs from \textit{Original} under every language setup ($R=1.40$, $1.35$ and $1.38$ on $\Delta z$ for \textit{Rephrase}, \textit{Empty} and \textit{Full}), commanding a slower lift each time ($-6.9$, $-9.3$ and $-9.2$\,mm/step, slower in $10/13$, $11/12$ and $11/13$ tasks). Since \textit{Rephrase} and \textit{Empty} leave the mass untouched and \textit{Empty} even asserts an empty container, this shift follows the reworded object name rather than the mass it implies. Under the vision-based cue, $\pi_0$-FAST and VLA-JEPA depart from \textit{Original} ($R=1.33$ and $1.15$), but they command a \emph{faster} lift for the heavier object ($+7.3$ and $+2.7$\,mm/step, faster in $10/10$ tasks), the opposite of the human pattern of lifting heavier objects more slowly~\cite{eastough2007}. Visible contents therefore change what these policies do without making them handle the added mass more carefully.

\section{Conclusion and Discussion}\label{sec:limitation}

We asked whether VLAs are aware of the physical properties of the objects they
manipulate, and whether they act on them. Across seven policies, three datasets and
twelve annotated properties, the answer to the first question is a qualified yes:
\emph{mass}, \emph{fragility}, \emph{deformability}, \emph{friction} and \emph{size}
are decodable from every modality stream,
but they are among the least prominent properties in the representation, and neither
robot pre-training nor downstream fine-tuning promotes them. The answer to the second is
no: when the same object is made heavier and the change is signalled in language or in
vision, the actions a policy commands for lifting move no further than
run-to-run variation on the original \emph{mass}, except where the surface form of the cue -- a reworded name or
visible contents -- moves them in a direction unrelated to the \emph{mass}.

The two halves of our study point to the same conclusion: physical properties are
weakly represented, and they are not connected to motion generation. One likely reason
is the training signal. Teleoperated demonstrations primarily capture whether a task
succeeds, rather than whether the motion handles each object safely and robustly.
Consequently, the collected data may favour a single motion strategy across many
objects, providing no signal for adapting motion to different physical properties.
Closing this gap requires data or objectives in which a physical property itself, rather
than the wording or appearance that accompanies it, changes the required motion.

\bibliographystyle{ieeetr}
\bibliography{references}

\end{document}